\documentclass[letterpaper]{article} 
\usepackage[preprint]{aaai2027}  
\usepackage[hyphens]{url}  
\usepackage{graphicx} 
\usepackage{natbib}  
\usepackage{caption} 
\usepackage{algorithm}
\usepackage{algorithmic}

\usepackage{newfloat}
\usepackage{listings}
\DeclareCaptionStyle{ruled}{labelfont=normalfont,labelsep=colon,strut=off} 
\floatstyle{ruled}
\newfloat{listing}{tb}{lst}{}
\floatname{listing}{Listing}

\usepackage{booktabs}
\usepackage{amsmath}
\usepackage{amssymb}
\usepackage{multirow}
\usepackage{bm}

\usepackage{booktabs}
\usepackage{multirow}
\usepackage{graphicx}
\usepackage{bm}
\usepackage{pifont}
\usepackage{xcolor}

\newcommand{\bestval}[1]{\ensuremath{\bm{#1}}}
\newcommand{\secondval}[1]{\ensuremath{\underline{#1}}}
\DeclareRobustCommand{\coarseblocklabel}{%
    \begingroup
    \setlength{\fboxsep}{1pt}%
    \colorbox[RGB]{218,226,245}{Coarse block}%
    \endgroup
}

\DeclareRobustCommand{\fineblocklabel}{%
    \begingroup
    \setlength{\fboxsep}{1pt}%
    \colorbox[RGB]{248,205,172}{Fine block}%
    \endgroup
}

\title{ProgFormer: Hierarchical Voxel Diffusion Transformer for Longitudinal Brain MRI Prediction}
\author{
    Dexuan Ding\textsuperscript{\rm 1}, 
    Yuankai Qi\textsuperscript{\rm 1}\corresponding,
    Luping Zhou\textsuperscript{\rm 2}, 
    Jian Yang\textsuperscript{\rm 1}, 
    Quan Z. Sheng\textsuperscript{\rm 1}, 
    Ming-Hsuan Yang\textsuperscript{\rm 3}
}
\affiliations{
    \textsuperscript{\rm 1}Macquarie University\\
    \textsuperscript{\rm 2}University of Sydney\\
    \textsuperscript{\rm 3}University of California Merced

    yuankai.qi@mq.edu.au
}

\begin{document}

\maketitle

\begin{abstract}
Predicting future structural  MRI of a brain is challenging because longitudinal changes are often subtle and 
confined to specific anatomical regions,
while most subject-specific brain structure remains stable over time. An effective model should therefore preserve global brain structural consistency while remaining sensitive to fine-grained disease progression. 
Existing latent-space-based methods improve computational efficiency, but suffer from information loss during their compression-reconstruction procedure.
In contrast, direct voxel-space methods avoid latent reconstruction but commonly use a unified prediction pathway to model brain structure and progression-related changes.
Subtle local changes may therefore be overshadowed by the dominant stable brain structure. 
To address these challenges, we propose ProgFormer, a hierarchical voxel-space Diffusion Transformer for longitudinal brain MRI prediction. 
ProgFormer uses a coarse pathway to perform the primary volumetric prediction from 3D patch tokens. This pathway models overall brain structure and longitudinal context. The fine pathway then uses the coarse representations as spatio-temporal grounding for voxel-level refinement within individual patches.
The two pathways jointly estimate a velocity field directly in voxel space through conditional flow matching, enabling end-to-end prediction without a separately learned image autoencoder. The predicted future scan is then generated from Gaussian noise by integrating the estimated velocity field over a sequence of Euler steps.
Extensive experimental results on three widely used benchmarks, ADNI, AIBL, and OASIS, under both pairwise and trajectory settings demonstrate favourable performance compared against several state-of-the-art methods.

\end{abstract}

\section{Introduction}

Predicting an individual’s future brain structural  MRI (sMRI) could facilitate the analysis and forecasting of Alzheimer’s disease progression~\cite{Jack2010, Aghajanian2025}. Unlike unconstrained image generation, longitudinal MRI prediction is strongly conditioned on the subject’s observed brain structure: most anatomical structures remain stable across visits, while disease progression manifests as localized changes in regions such as the hippocampus, lateral ventricles, and surrounding cortex~\cite{Frisoni2010, Wang2026}. These changes occur at multiple spatial scales, ranging from coherent structural alterations, such as ventricular enlargement, to subtle boundary displacements and voxel-level intensity variations associated with hippocampal and cortical atrophy. Accurate prediction should therefore preserve subject-specific brain structure while sensitively resolving localized progression-related changes at voxel resolution~\cite{Verdi, 10.1162/imag_a_00294, Ledig2018}.

\begin{figure}[!tbp]
    \centering
    \includegraphics[
        width=\linewidth
    ]{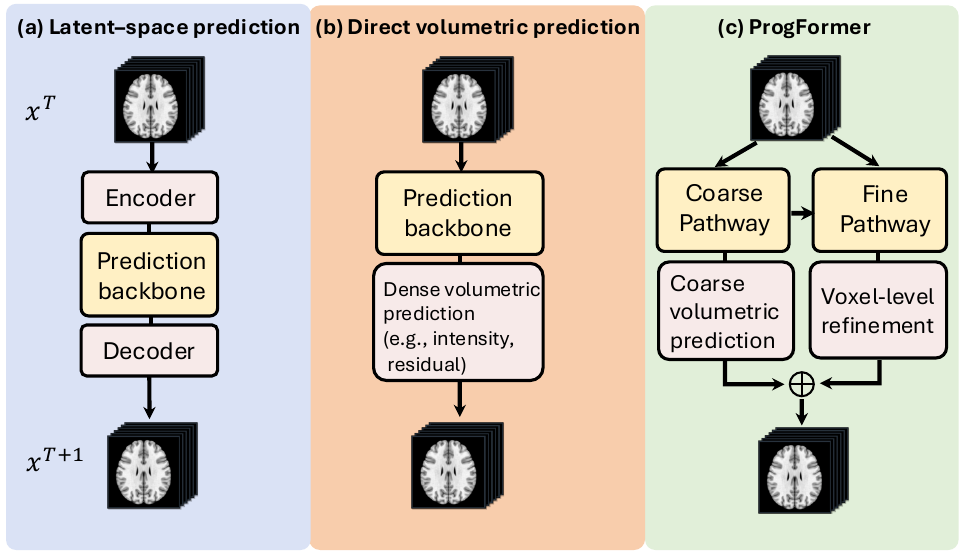}
    \caption {Longitudinal brain MRI prediction paradigms.
    (a) Latent-space methods encode MRI scan before prediction and decode the predicted representation back to voxel space. 
    (b) Direct volumetric methods use a unified backbone to produce a dense prediction, requiring the same backbone to model brain structure and progression-related changes across spatial scales.
    (c) ProgFormer retains a coarse pathway that performs the primary volumetric prediction at the patch level, serving a role similar to the unified backbone in (b). A fine pathway then uses the coarse patch representations as spatio-temporal grounding for voxel-level refinement.}
    \label{fig:motivation}
\end{figure}

\begin{figure*}[tbp]
    \centering
    \includegraphics[
        width=1.\linewidth
    ]{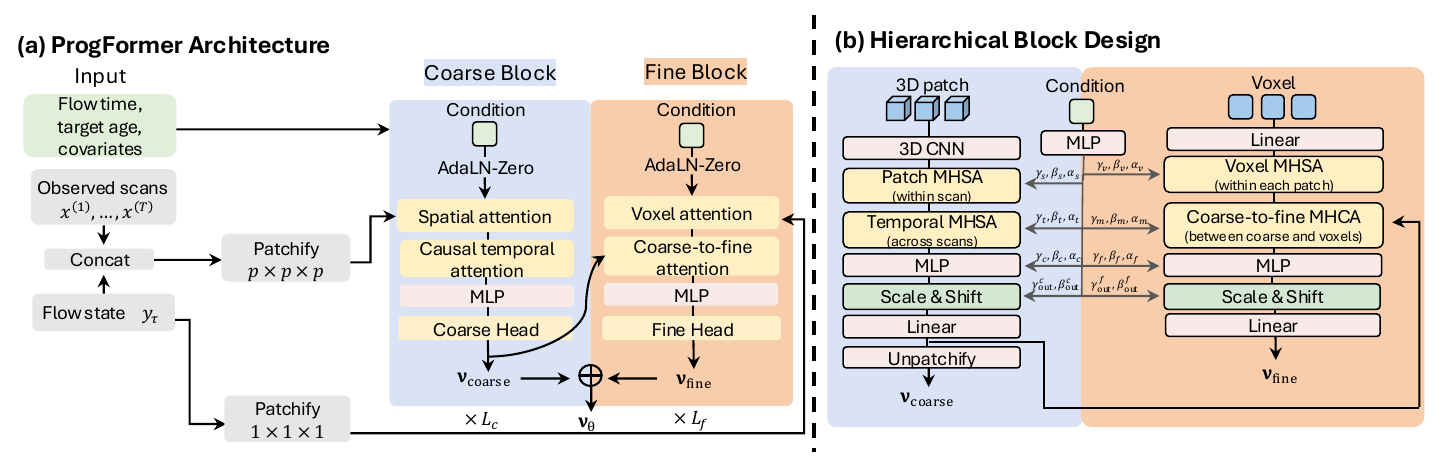}
    \caption{Overview of ProgFormer. (a) ProgFormer processes the observed scans and current flow state through a coarse patch pathway and a coarse-guided voxel pathway. The coarse and fine velocities are added to form the predicted velocity $\mathbf{\nu}_{\theta}$. (b) Each \coarseblocklabel{} applies spatial and causal temporal MHSA to patch tokens, while each \fineblocklabel{} applies voxel MHSA and coarse-to-fine MHCA within each patch.}
    \label{fig:progformer}
\end{figure*}
\begin{figure}[t]
    \centering
    \includegraphics[
        width=1.\linewidth
    ]{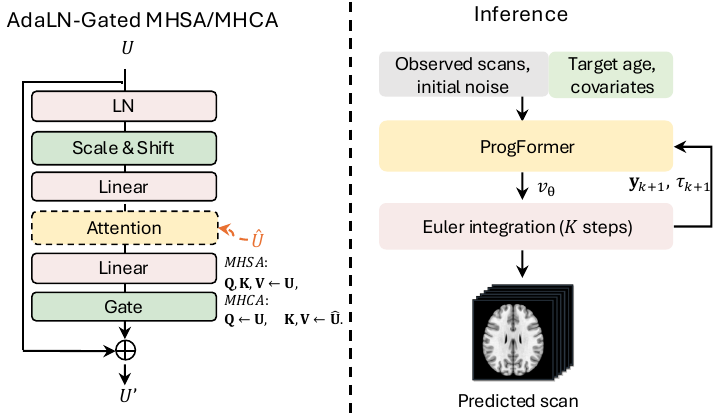}
    \caption{AdaLN-gated attention and Euler inference. Left: MHSA obtains queries, keys, and values from $\mathbf{U}$, whereas MHCA obtains queries from $\mathbf{U}$ and keys and values from $\widehat{\mathbf{U}}$. Conditioning modulates and gates the attention update. Right: Starting from Gaussian noise, ProgFormer predicts a velocity and updates the flow state for $K$ Euler steps. The final state is the predicted future MRI.}
    \label{fig:adaln_attention}
    \vspace{-0.4cm}
\end{figure}

Existing approaches
roughly fall into two groups:
latent-space-based or direct voxel-space prediction.
As illustrated in Figure~\ref{fig:motivation}(a), latent-space methods encode MRI scans into low-dimensional representations, predict the future representation in latent space, and decode it back to voxel space~\cite{BrLP,AG-LDM,DeltaLFM}. 
This reduces the dimensionality processed by the prediction backbone 
but introduces a separately learned autoencoder. The fidelity of the final voxel-level prediction therefore also depends on how well the encoding and decoding stages preserve fine structural information.
which may be affected by the compression
~\cite{MedVAE,WaveletFusionMRI}.
%
%
Direct volumetric approaches instead predict outputs directly in voxel space.
They are able to produce outputs such as a complete follow-up MRI, a voxel-wise intensity residual, or a dense deformation field applied to the observed scan~\cite{SADM,TADM3D,countersynth}. 
Despite these different formulations, the dense prediction or update is predicted through a unified output pathway. 
This pathway has to capture subject-specific whole-brain structure while resolving progression-related changes across different spatial scales. Since most brain structure remains stable between doctor's visits, full-image targets are dominated by stable content, while residual and deformation targets contain predominantly small voxel-wise updates~\cite{Frisoni2010,DeltaLFM}. Subtle boundary displacements and intensity variations can therefore be difficult to resolve when the same pathway is responsible for both overall brain structure and localised progression-related changes.

%
Recent advances in pixel-space Diffusion Transformers introduce new architectures for modelling images directly without a learned autoencoder. PixelDiT adopts a dual-level design in which a patch-level DiT captures global semantics and a pixel-level DiT refines local details~\cite{PixelDiT}. JiT follows a different approach, using plain Transformers to predict clean images directly in pixel space~\cite{JiT}. DiT-3D and PointDiT further extend Transformer-based generative modeling to 3D point clouds and dense point maps, respectively~\cite{DiT3D,PointDiT}. These studies demonstrate the potential of direct Transformer-based modelling beyond latent representations. 
However, it remains unknown whether the above direct transformer modelling paradigm applies to voxel-space MRI prediction, where the architecture should preserve brain structure while resolving subtle progression-related changes. Unlike the generation settings considered above, longitudinal MRI prediction also requires temporal reasoning over observed scans.


In this paper, we introduce \textbf{ProgFormer}, a hierarchical voxel-space diffusion transformer
for longitudinal brain MRI prediction. 
%
As shown in Figure~\ref{fig:motivation}(c), ProgFormer structures voxel-space prediction into a coarse pathway and a fine pathway. The coarse pathway processes 3D patch tokens, using spatial attention within each scan and causal temporal attention across observed scans at corresponding patch locations to model brain structure, patch-scale changes, and longitudinal context. The fine pathway applies voxel attention within each 3D patch and uses the corresponding coarse representation as spatio-temporal grounding for voxel-level refinement of local structural details and intensity variations.
The two pathways jointly predict a voxel-space velocity field. Starting from Gaussian noise, ProgFormer integrates this field over a sequence of Euler steps and returns the final flow state as the predicted future MRI.~\cite{lipman2023flow,DeltaLFM}.

We evaluate ProgFormer under two settings. The pairwise setting 
requires predicting one future scan given one already observed scan.
It covers different starting visits and follow-up intervals.
The trajectory setting requires predicting the final scan given a sequence of previous scans.
Together, they assess prediction across varying longitudinal intervals and the effectiveness of observed history for a fixed final target.

Our main contributions are summarized as follows:
\begin{itemize}
    \item We present ProgFormer, an end-to-end hierarchical voxel-space diffusion transformer for longitudinal 3D brain MRI prediction without a separately trained image autoencoder.

    \item We design a new hierarchical voxel-space diffusion paradigm for longitudinal 3D MRI prediction, introducing the first unified architecture to jointly perform global patch-level spatio-temporal reasoning and fine-grained intra-patch voxel modeling for anatomically faithful, progression-sensitive forecasting.

    \item Extensive experimental results on three widely adopted benchmarks, ADNI, AIBL, and OASIS, demonstrate favourable performance compared against several state-of-the-art methods.
\end{itemize}

\section{Related Work}
\paragraph{Longitudinal brain MRI prediction.}
Recent methods generate follow-up MRI from baseline scans, clinical variables, or observed scans. BrLP and AG-LDM employ latent diffusion with disease-specific priors or anatomical supervision, while $\Delta$-LFM learns patient-specific progression trajectories through latent flow matching~\cite{BrLP,AG-LDM,DeltaLFM}. TADM-3D directly predicts volumetric progression with temporal regularisation, whereas SADM introduces sequence-aware conditioning for longitudinal generation~\cite{TADM3D,SADM}. Collectively, these methods condition future-image generation using temporal, clinical, or structural information. ProgFormer instead studies how patch-level spatiotemporal modeling and voxel-level refinement can be coupled for direct volumetric prediction.

\paragraph{Pixel-space Diffusion Transformers.}
Pixel-space Diffusion Transformers generate images without a separately trained autoencoder. PixelDiT combines patch-level modeling with pixel-level refinement, while JiT demonstrates direct image generation using large-patch Transformers~\cite{PixelDiT,JiT}. Related 3D models include DiT-3D, which operates on voxelized point clouds, and PointDiT, which predicts dense point maps directly in pixel space~\cite{DiT3D,PointDiT}. ProgFormer adapts this direction to longitudinal brain MRI by introducing a patch-to-voxel hierarchy with spatial and temporal reasoning over observed scans.





\section{Method}
\label{sec:method}

\subsection{Task Formulation}
\label{sec:task}

Let $T$ denote the number of observed scans used for a prediction, indexed chronologically by $i\in\{1,\ldots,T\}$. The corresponding structural MRI, acquisition age, and diagnosis at index $i$ are denoted by $\mathbf{x}^{(i)}\in\mathbb{R}^{D\times H\times W}$, $a^{(i)}$, and $g^{(i)}$, where $D$, $H$, and $W$ denote the depth, height, and width of the scan. Let $\Omega=\{1,\ldots,D\}\times\{1,\ldots,H\}\times\{1,\ldots,W\}$ denote its discrete spatial domain. Each coordinate $(d,h,w)\in\Omega$ identifies one voxel, whose intensity in $\mathbf{x}^{(i)}$ is denoted by $x^{(i)}_{d,h,w}\in\mathbb{R}$. Sex is denoted by $\xi$. The target scan and its acquisition age are denoted by $\mathbf{x}^{(T+1)}$ and $a^{(T+1)}$, respectively.

We denote the age interval between the last acquisition age of $\mathbf{x}^{(T)}$ and the target age by $\Delta a_T=a^{(T+1)}-a^{(T)}$.

The observed longitudinal history and covariate vector are defined as
\begin{equation}
    \begin{aligned}
        \mathcal{H}_T
        &=
        (
            \mathbf{x}^{(i)},a^{(i)}
        )_{i=1}^{T},
        \qquad
        a^{(1)}<\cdots<a^{(T)},\\
        \mathbf{c}_T
        &=
        [
            a^{(T)},
            \Delta a_T,
            g^{(T)},
            \xi
        ].
    \end{aligned}
    \label{eq:history}
\end{equation}
Given $\mathcal{H}_T$ and $\mathbf{c}_T$, the task is to predict $\mathbf{x}^{(T+1)}$ at target age $a^{(T+1)}$.

\subsection{Preliminaries}
\label{sec:preliminaries}

\paragraph{Conditional flow matching.}
Flow matching learns a continuous velocity field that transports samples from a source distribution to a target distribution~\cite{lipman2023flow}.In our task, the source is a standard Gaussian distribution, while the target is the conditional data distribution of ground-truth future MRI scans given the observed history, target age, and covariates. Let $\tau\in[0,1]$ denote flow time, which indexes the progress of generation from the source at $\tau=0$ to the target at $\tau=1$. 

Let $\mathbf{y}_{\tau}$ denote the flow state at flow time $\tau$. ProgFormer learns the conditional velocity field
\begin{equation}
    \frac{\mathrm{d}\mathbf{y}_{\tau}}{\mathrm{d}\tau}
    =
    \mathbf{\nu}_{\theta}
    \left(
        \mathbf{y}_{\tau},\tau
        \mid
        \mathcal{H}_T,
        a^{(T+1)},
        \mathbf{c}_T
    \right),
    \label{eq:flow_ode}
\end{equation}
where $\theta$ denotes the learnable parameters of ProgFormer. The velocity field specifies the instantaneous change of the flow state along the generative path. Integrating Eq.~\eqref{eq:flow_ode} from $\tau=0$ to $\tau=1$ transforms a Gaussian noise sample into a predicted future sMRI.

During training,  $\mathbf{y}_0$ is set to $\boldsymbol{\epsilon}$, where $\boldsymbol{\epsilon}\sim\mathcal{N}(\mathbf{0},\mathbf{I})$, and let $\mathbf{y}_1=\mathbf{x}^{(T+1)}$ be the ground-truth target scan. For $\tau\sim\mathcal{U}(0,1)$, we construct the linear interpolation path
\begin{equation}
    \mathbf{y}_{\tau}
    =
    (1-\tau)\mathbf{y}_0+\tau\mathbf{y}_1.
    \label{eq:flow_path}
\end{equation}
The velocity associated with this path is
\begin{equation}
    \mathbf{\nu}^{*}
    =
    \frac{\mathrm{d}\mathbf{y}_{\tau}}{\mathrm{d}\tau}
    =
    \mathbf{y}_1-\mathbf{y}_0.
    \label{eq:target_velocity}
\end{equation}
For each sampled noise-target pair and flow time, $\mathbf{\nu}^{*}$ is the supervision target for the velocity predicted by ProgFormer at $\mathbf{y}_{\tau}$. The corresponding 
flow-matching objective is described in Section~\ref{sec:training}.

\paragraph{Conditioning.}
The flow time, target age, and covariate vector are embedded independently and combined by element-wise addition:
\begin{equation}
    \mathbf{h}
    =
    E_{\tau}(\tau)
    +
    E_a\!(a^{(T+1)})
    +
    E_c(\mathbf{c}_T),
    \label{eq:conditioning}
\end{equation}
where $E_{\tau}$, $E_a$, and $E_c$ are learnable embedding functions whose outputs have the same dimension. The resulting vector $\mathbf{h}$ is shared across all tokens, allowing the attention and MLP modules to adapt their computations to the current flow time, target age, and covariates. Following DiT~\cite{DiT}, we inject this global conditioning through adaptive layer normalization with zero initialization (AdaLN-Zero).

Let $r$ index an attention or MLP module, and let $F_r$ denote the computation performed by that module. The indexed module can be a layer in either the coarse pathway described in Section~\ref{sec:coarse_pathway} or the fine pathway described in Section~\ref{sec:fine_pathway}.
For input 
$\mathbf{U}$, a modulation MLP maps $\mathbf{h}$ to a scale $\gamma_r(\mathbf{h})$, shift $\beta_r(\mathbf{h})$, and gate $\alpha_r(\mathbf{h})$. The scale and shift modify the normalized token features, while the gate controls how much of the transformed features is added to the input:
\small{
\begin{align}
\operatorname{AdaLN}_r(\mathbf{U},\mathbf{h})
    &=
    \left(
        1+\gamma_r(\mathbf{h})
    \right)
    \odot
    \operatorname{LN}(\mathbf{U})
    +
    \beta_r(\mathbf{h}),
    \nonumber\\
    \mathbf{U}'
    &=
    \mathbf{U}
    +
    \alpha_r(\mathbf{h})
    \odot
    F_r\!\left(
        \operatorname{AdaLN}_r(\mathbf{U},\mathbf{h})
    \right).
    \label{eq:adaln_update}
\end{align}
}
The operator $\operatorname{LN}$ denotes layer normalization. The gate scales the output of $F_r$ before it is added to the input tokens. 

\subsection{Method Overview}
\label{sec:overview}

As shown in Figure~\ref{fig:progformer}, ProgFormer processes the observed scans together with the flow state at each flow time. The coarse pathway applies spatial and causal temporal attention to 3D patch tokens, producing the primary patch-level velocity and coarse representations that encode whole-brain structure and longitudinal information. 
The fine pathway operates on voxel tokens from the flow state and uses the corresponding coarse patch tokens, which encode whole-brain and longitudinal information, to guide voxel-level refinement.
The coarse and fine velocities are added to form the final velocity predicted by ProgFormer. During training, this prediction is supervised by the target velocity defined in Section~\ref{sec:preliminaries}. During inference, ProgFormer integrates the predicted velocity field from Gaussian noise over a sequence of Euler steps to generate the future MRI, which we discuss in Section~\ref{sec:training}. 

\subsection{Coarse Pathway}
\label{sec:coarse_pathway}

\paragraph{Input and patch embedding.}
The coarse pathway processes the observed scans together with the current flow state $\mathbf{y}_{\tau}$. During training, $\mathbf{y}_{\tau}$ is constructed using the interpolation path in Eq.~\eqref{eq:flow_path}. During inference, it is the state obtained at the current Euler step, as described in Section~\ref{sec:training}. The observed scans and flow state are combined as:
$
    \mathbf{X}_{\tau}
    =
    [
        \mathbf{x}^{(1)},\ldots,\mathbf{x}^{(T)},
        \mathbf{y}_{\tau}
    ]
    \in
    \mathbb{R}^{(T+1)\times D\times H\times W}.
    \label{eq:model_sequence}
$

Each input in $\mathbf{X}_{\tau}$ is embedded as non-overlapping $p\times p\times p$ patch tokens using a 3D convolution with kernel size and stride $p$. Let $N=(D/p)\times(H/p)\times(W/p)$ denote the number of patches per input and let $d_c$ denote the coarse-token dimension. The token at patch location $n$ in input $i$ is denoted by $\Phi_p(\mathbf{X}_{\tau,i})_n\in\mathbb{R}^{d_c}$.

Spatial-position and acquisition-age embeddings are added to each patch token:
\begin{equation}
    \mathbf{z}_{i,n}
    =
    \Phi_p(\mathbf{X}_{\tau,i})_n
    +
    \mathbf{e}^{\mathrm{pos}}_n
    +
    \mathbf{e}^{\mathrm{age}}(a^{(i)}),
    \qquad
    i\in\{1,\ldots,T+1\}.
    \label{eq:coarse_token_embedding}
\end{equation}
For $i\leq T$, $a^{(i)}$ is the acquisition age of the corresponding observed scan $\mathbf{x}^{(i)}$. For $i=T+1$, the target age $a^{(T+1)}$ is assigned to the flow state $\mathbf{y}_{\tau}$. Collecting all patch tokens gives
$
    \mathbf{Z}
    \in
    \mathbb{R}^{(T+1)\times N\times d_c}.
    \label{eq:coarse_tokens}
$    

\paragraph{Coarse block.} 
As shown in Figure~\ref{fig:progformer}(b), each coarse block takes patch tokens $\mathbf{Z}$ as input.
The spatial MHSA follows the AdaLN-gated structure illustrated in Figure~\ref{fig:adaln_attention} and uses the AdaLN-Zero conditioning described in Section~\ref{sec:preliminaries}. Using the subscript $s$ to denote spatial attention, the resulting tokens are
\begin{equation}
    \widetilde{\mathbf{Z}}
    =
    \mathbf{Z}
    +
    \alpha_s(\mathbf{h})
    \odot
    \operatorname{MHSA}_s
    \left(
        \operatorname{AdaLN}_s(\mathbf{Z},\mathbf{h})
    \right).
    \label{eq:coarse_spatial}
\end{equation}

The block next applies causal temporal MHSA across the $T+1$ inputs at each patch location. At location $n$, it processes $[\widetilde{\mathbf{z}}_{1,n},\ldots,\widetilde{\mathbf{z}}_{T,n},\widetilde{\mathbf{z}}_{T+1,n}]$. Following masked self-attention~\cite{Vaswani2017}, we apply a lower-triangular causal mask so that each token can attend only to itself and earlier inputs. Because the flow state occupies the final position, $\widetilde{\mathbf{z}}_{T+1,n}$ can attend to the spatially updated tokens from all observed scans at that location, while the observed-scan tokens cannot attend to the flow state.

To support different numbers of observed scans, shorter input sequences are zero-padded to a common length. A padding mask prevents the padded tokens from being used as keys or values in temporal MHSA. The padding and causal masks are combined to form $\mathbf{M}$. The temporal MHSA uses the same AdaLN-gated structure and conditioning mechanism. Using the subscript $t$ to denote temporal attention, the resulting tokens are
\begin{equation}
    \overline{\mathbf{Z}}
    =
    \widetilde{\mathbf{Z}}
    +
    \alpha_t(\mathbf{h})
    \odot
    \operatorname{MHSA}_t
    \left(
        \operatorname{AdaLN}_t(\widetilde{\mathbf{Z}},\mathbf{h}),
        \mathbf{M}
    \right).
    \label{eq:coarse_temporal}
\end{equation}

Finally, the conditioned multilayer perceptron (MLP) transforms the spatially and temporally updated patch representations. Using the subscript $c$ to denote the coarse MLP, the block output is
\begin{equation}
    \mathbf{Z}'
    =
    \overline{\mathbf{Z}}
    +
    \alpha_c(\mathbf{h})
    \odot
    \operatorname{MLP}_c
    \left(
        \operatorname{AdaLN}_c(\overline{\mathbf{Z}},\mathbf{h})
    \right).
    \label{eq:coarse_mlp}
\end{equation}

After $L_c$ coarse blocks, we retain the $N$ output tokens corresponding to the flow state $\mathbf{y}_{\tau}$ and denote them by $\mathbf{Z}_{\tau}\in\mathbb{R}^{N\times d_c}$.

A final conditioned normalization transforms the retained flow-state tokens $\mathbf{Z}_{\tau}$ into the coarse representations $\mathbf{C}$ used by the coarse prediction head and provided to the fine pathway as spatial and longitudinal context for voxel-level refinement:
\begin{equation}
    \mathbf{C}
    =
    \operatorname{AdaLN}^{c}_{\mathrm{out}}
    \left(
        \mathbf{Z}_{\tau},
        \mathbf{h}
    \right)
    \in
    \mathbb{R}^{N\times d_c}.
    \label{eq:coarse_features}
\end{equation}

\paragraph{Coarse prediction head.}
Let $P=p^3$ denote the number of voxels in each patch. A linear layer maps each row of $\mathbf{C}$ to $P$ velocity values. These values are shaped into $p\times p\times p$ blocks and placed at their 3D patch locations to form
$
    \mathbf{\nu}_{\mathrm{coarse}}
    \in
    \mathbb{R}^{D\times H\times W}.
    \label{eq:coarse_prediction}
$    

\subsection{Fine Pathway}
\label{sec:fine_pathway}

The fine pathway represents the individual voxel intensities of the flow state as tokens and uses the coarse representations as spatial and longitudinal context for voxel-level refinement.

\paragraph{Voxel embedding.}
The flow state $\mathbf{y}_{\tau}$ is divided into $N$ non-overlapping $p\times p\times p$ patches, each containing $P=p^3$ voxels. Let $y_{\tau,n,j}\in\mathbb{R}$ denote the intensity of voxel $j$ in patch $n$, where $n\in\{1,\ldots,N\}$ and $j\in\{1,\ldots,P\}$. The coarse representation associated with patch $n$ is projected to the voxel-token dimension:
\begin{equation}
    \widehat{\mathbf{C}}_n
    =
    \mathbf{C}_n\mathbf{W}_{cf}
    +
    \mathbf{b}_{cf}
    \in
    \mathbb{R}^{d_f},
    \qquad
    \mathbf{W}_{cf}
    \in
    \mathbb{R}^{d_c\times d_f}.
    \label{eq:coarse_fine_projection}
\end{equation}
Each voxel intensity is independently mapped to a $d_f$-dimensional token using a linear layer, equivalent to a $1\times1\times1$ patch embedding, and combined with its projected coarse representation:
\begin{equation}
    \mathbf{v}_{n,j}
    =
    y_{\tau,n,j}\mathbf{w}_v
    +
    \mathbf{b}_v
    +
    \widehat{\mathbf{C}}_n,
    \qquad
    \mathbf{w}_v,\mathbf{b}_v
    \in
    \mathbb{R}^{d_f}.
    \label{eq:voxel_tokenization}
\end{equation}
Collecting the voxel tokens and projected coarse representations gives
\begin{equation}
    \mathbf{V}
    \in
    \mathbb{R}^{N\times P\times d_f},
    \qquad
    \widehat{\mathbf{C}}
    \in
    \mathbb{R}^{N\times 1\times d_f}.
    \label{eq:fine_tokens}
\end{equation}

\paragraph{Fine block.}
Let $\mathbf{V}\in\mathbb{R}^{N\times P\times d_f}$ denote the voxel-token entering a fine block. For the first fine block, $\mathbf{V}$ is obtained from the voxel embedding in Eq.~\eqref{eq:fine_tokens}. The voxel MHSA and coarse-to-fine multi-head cross-attention (MHCA) follow the respective AdaLN-gated structures illustrated in Figure~\ref{fig:adaln_attention} and use the AdaLN-Zero conditioning described in Section~\ref{sec:preliminaries}. Using the subscript $v$ to denote voxel MHSA, the resulting tokens are
\begin{equation}
    \widetilde{\mathbf{V}}
    =
    \mathbf{V}
    +
    \alpha_v(\mathbf{h})
    \odot
    \operatorname{MHSA}_v
    \left(
        \operatorname{AdaLN}_v(\mathbf{V},\mathbf{h})
    \right).
    \label{eq:fine_voxel}
\end{equation}

Coarse-to-fine MHCA then incorporates the coarse representation associated with each patch. For patch $n$, the voxel tokens provide the queries, while $\widehat{\mathbf{C}}_n$ provides the key and value: $\mathbf{Q}^{cf}_n\leftarrow\widetilde{\mathbf{V}}_n$ and $\mathbf{K}^{cf}_n,\mathbf{V}^{cf}_n\leftarrow\widehat{\mathbf{C}}_n$. This gives the voxel tokens access to the spatial and longitudinal information encoded by the coarse pathway. Using the subscript $m$ to denote coarse-to-fine MHCA, the resulting coarse-guided voxel tokens are
\begin{equation}
    \overline{\mathbf{V}}
    =
    \widetilde{\mathbf{V}}
    +
    \alpha_m(\mathbf{h})
    \odot
    \operatorname{MHCA}_m
    \left(
        \operatorname{AdaLN}_m(\widetilde{\mathbf{V}},\mathbf{h}),
        \widehat{\mathbf{C}}
    \right).
    \label{eq:fine_cross}
\end{equation}

The conditioned multilayer perceptron (MLP) subsequently transforms the coarse-guided voxel representations. Using the subscript $f$ to denote the fine MLP, the block output is
\begin{equation}
    \mathbf{V}'
    =
    \overline{\mathbf{V}}
    +
    \alpha_f(\mathbf{h})
    \odot
    \operatorname{MLP}_f
    \left(
        \operatorname{AdaLN}_f(\overline{\mathbf{V}},\mathbf{h})
    \right).
    \label{eq:fine_mlp}
\end{equation}
The output $\mathbf{V}'$ is used as the input $\mathbf{V}$ to the next fine block. After applying $L_f$ fine blocks, the resulting output is denoted by $\mathbf{V}_{\mathrm{out}}\in\mathbb{R}^{N\times P\times d_f}$.

\paragraph{Fine prediction head.}
The fine prediction head applies a final conditioned normalization:
\begin{equation}
    \mathbf{F}
    =
    \operatorname{AdaLN}^{f}_{\mathrm{out}}
    \left(
        \mathbf{V}_{\mathrm{out}},
        \mathbf{h}
    \right)
    \in
    \mathbb{R}^{N\times P\times d_f}.
    \label{eq:fine_features}
\end{equation}
A linear layer maps each voxel token to one velocity value, producing
\begin{equation}
    \mathbf{V}_{\mathrm{fine}}
    =
    \mathbf{F}\mathbf{W}_f+\mathbf{b}_f
    \in
    \mathbb{R}^{N\times P},
    \qquad
    \mathbf{W}_f
    \in
    \mathbb{R}^{d_f\times 1}.
    \label{eq:fine_patch_prediction}
\end{equation}
For each patch, its $P$ values are shaped into a $p\times p\times p$ block and placed at their corresponding voxel locations to form
$
    \mathbf{\nu}_{\mathrm{fine}}
    \in
    \mathbb{R}^{D\times H\times W}.
    \label{eq:fine_prediction}
$
The complete predicted velocity is
\begin{equation}
    \mathbf{\nu}_{\theta}
    =
    \mathbf{\nu}_{\mathrm{coarse}}
    +
    \mathbf{\nu}_{\mathrm{fine}},
    \label{eq:velocity_decomposition}
\end{equation}
where the coarse pathway provides the primary voxel-space velocity and the fine pathway provides coarse-guided voxel-level refinement.

\begin{table*}[!ht]
\centering
\caption{
Results comparison against several state-of-the-art methods under two settings.
Best results are highlighted in \textbf{bold}, and second-best results are \underline{underlined}. SSIM is presented on a $0$--$100$ scale.}
\label{tab:main_results}
\scriptsize
\renewcommand{\arraystretch}{1.16}
\setlength{\tabcolsep}{2.6pt}
\resizebox{\textwidth}{!}{
\begin{tabular}{lclccc@{\hspace{7pt}}ccc@{\hspace{7pt}}ccc}
\toprule
\multirow[c]{2}{*}{\textbf{Protocol}} & \multirow[c]{2}{*}{\textbf{Input scans (${I}$)}} & \multirow[c]{2}{*}{\textbf{Method}} & \multicolumn{3}{c}{\textbf{ADNI}} & \multicolumn{3}{c}{\textbf{AIBL}} & \multicolumn{3}{c}{\textbf{OASIS}} \\
\cmidrule(lr){4-6}
\cmidrule(lr){7-9}
\cmidrule(lr){10-12}
&&& \textbf{PSNR} $\uparrow$ & \textbf{SSIM} $\uparrow$ & \textbf{R-MAE} $\downarrow$ & \textbf{PSNR} $\uparrow$ & \textbf{SSIM} $\uparrow$ & \textbf{R-MAE} $\downarrow$ & \textbf{PSNR} $\uparrow$ & \textbf{SSIM} $\uparrow$ & \textbf{R-MAE} $\downarrow$ \\
\midrule
\multirow[c]{6}{*}{Pairwise} & \multirow[c]{6}{*}{$1$} & BrLP~\cite{BrLP} & $25.89 \pm 2.33$ & $92.97 \pm 1.83$ & $0.280 \pm 0.163$ & $25.65 \pm 2.52$ & $92.92 \pm 2.15$ & $0.284 \pm 0.170$ & $25.52 \pm 2.31$ & $92.99 \pm 1.57$ & $0.313 \pm 0.232$ \\
&& CounterSynth~\cite{countersynth} & \secondval{26.19 \pm 2.64} & \secondval{93.71 \pm 1.93} & \secondval{0.234 \pm 0.117} & $26.73 \pm 3.83$ & \secondval{93.73 \pm 3.12} & $0.251 \pm 0.133$ & \secondval{26.04 \pm 2.43} & \secondval{93.68 \pm 2.08} & \secondval{0.284 \pm 0.133} \\
&& TADM-3D~\cite{TADM3D} & $25.79 \pm 1.35$ & $92.69 \pm 2.29$ & $0.311 \pm 0.187$ & $23.12 \pm 1.34$ & $90.95 \pm 2.45$ & $0.256 \pm 0.163$ & $24.90 \pm 1.88$ & $93.35 \pm 2.22$ & $0.320 \pm 0.196$ \\
&& SADM~\cite{SADM} & $24.60 \pm 1.79$ & $91.33 \pm 2.13$ & $0.441 \pm 0.302$ & $21.96 \pm 1.74$ & $83.41 \pm 3.37$ & $0.389 \pm 0.265$ & $24.11 \pm 1.22$ & $90.05 \pm 3.13$ & $0.408 \pm 0.286$ \\
&& ProgFormer-S (Ours) & \bestval{26.39 \pm 1.68} & $93.56 \pm 2.12$ & \secondval{0.234 \pm 0.138} & \bestval{26.95 \pm 2.52} & \bestval{93.91 \pm 2.73} & \secondval{0.245 \pm 0.125} & \bestval{26.32 \pm 1.54} & \bestval{94.20 \pm 2.11} & $0.303 \pm 0.163$ \\
&& ProgFormer-I (Ours) & $26.11 \pm 1.76$ & \bestval{94.34 \pm 1.60} & \bestval{0.230 \pm 0.135} & \secondval{26.79 \pm 2.37} & $92.42 \pm 3.87$ & \bestval{0.223 \pm 0.161} & $25.36 \pm 1.58$ & $93.08 \pm 2.60$ & \bestval{0.281 \pm 0.157} \\
\midrule
\midrule
\multirow[c]{5}{*}{Trajectory} & \multirow[c]{3}{*}{$1$} & BrLP~\cite{BrLP} & $26.62 \pm 2.14$ & $93.94 \pm 1.92$ & $0.289 \pm 0.175$ & $25.94 \pm 2.21$ & $93.06 \pm 2.25$ & $0.359 \pm 0.167$ & $25.97 \pm 2.29$ & $93.14 \pm 1.89$ & $0.353 \pm 0.247$ \\
&& CounterSynth~\cite{countersynth} & $26.08 \pm 2.62$ & \secondval{95.80 \pm 1.92} & \secondval{0.236 \pm 0.124} & \secondval{26.79 \pm 3.89} & \secondval{93.64 \pm 3.20} & \secondval{0.334 \pm 0.192} & \secondval{26.38 \pm 2.25} & \bestval{93.51 \pm 1.77} & \bestval{0.342 \pm 0.204} \\
&& TADM-3D~\cite{TADM3D} & $23.35 \pm 1.97$ & $92.78 \pm 2.24$ & $0.313 \pm 0.198$ & $23.53 \pm 1.37$ & $91.50 \pm 2.72$ & $0.355 \pm 0.221$ & $24.89 \pm 1.55$ & $93.37 \pm 2.94$ & $0.371 \pm 0.216$ \\
\cmidrule(lr){2-12}
& \multirow[c]{2}{*}{$T$} & SADM~\cite{SADM} & \secondval{27.32 \pm 2.39} & $92.08 \pm 3.49$ & $0.355 \pm 0.251$ & $24.16 \pm 1.95$ & $85.22 \pm 3.24$ & $0.370 \pm 0.227$ & $25.21 \pm 1.47$ & $91.23 \pm 3.14$ & $0.370 \pm 0.244$ \\
&& ProgFormer-S (Ours) & \bestval{27.90 \pm 1.63} & \bestval{96.17 \pm 2.04} & \bestval{0.192 \pm 0.118} & \bestval{26.96 \pm 2.51} & \bestval{93.84 \pm 2.71} & \bestval{0.321 \pm 0.173} & \bestval{27.17 \pm 1.50} & \secondval{93.45 \pm 2.48} & \secondval{0.344 \pm 0.253} \\
\bottomrule
\end{tabular}
}
\end{table*}

\subsection{Sequential and Interleaved Compositions}
\label{sec:variants}


We evaluate sequential (ProgFormer-S) and interleaved (ProgFormer-I) compositions. ProgFormer-S completes all coarse blocks before voxel refinement, with the final coarse representations supplied to every fine block. 



\begin{figure}[tbp]
    \centering
    \includegraphics[
        width=0.47\textwidth
    ]{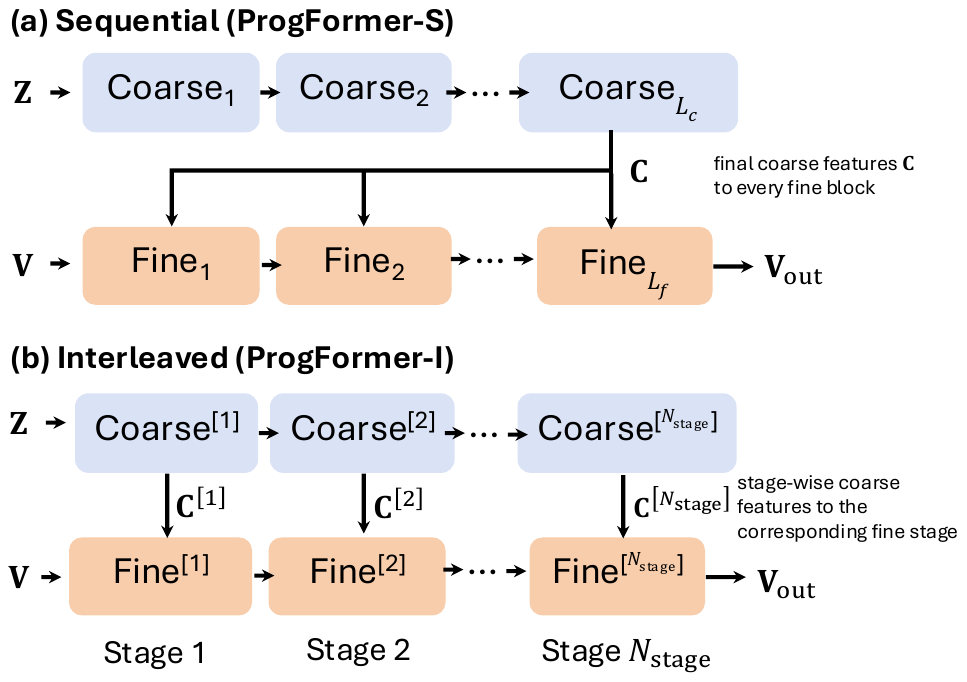}
    \caption{\textbf{Sequential and interleaved pathway composition.}
    }
    \label{fig:variants}
\end{figure}

\paragraph{Sequential variant (ProgFormer-S).}

The sequential variant applies all $L_c$ coarse blocks before the $L_f$
fine blocks. Every fine block receives the final coarse features
$\mathbf{C}$. Voxel-level refinement therefore begins after patch-level
spatial and temporal modeling is complete.

\paragraph{Interleaved variant (ProgFormer-I).}


The interleaved variant divides both pathways into $N_{\mathrm{stage}}$ stages. Each stage contains $L_c/N_{\mathrm{stage}}$ coarse blocks followed by $L_f/N_{\mathrm{stage}}$ fine blocks. At stage $\ell$, the fine blocks receive the target-state coarse features produced by coarse stage $\ell$. Early fine blocks therefore use shallower coarse representations, while later blocks receive progressively refined spatial and temporal context. Information flows from coarse to fine without feeding fine features back into subsequent coarse stages.

The sequential variant transfers the final coarse representation throughout
voxel refinement. The interleaved variant instead distributes coarse-to-fine
transfer across network depth. We use the sequential variant as the primary
model and compare both compositions experimentally.

\subsection{Training and Inference}
\label{sec:training}

\paragraph{Training objective.}
Before training, SynthSeg~\cite{Billot2023SynthSeg} is applied to each ground-truth future scan $\mathbf{x}^{(T+1)}$ to obtain an anatomical label map $\mathbf{S}^{(T+1)}\in\mathcal{R}^{D\times H\times W}$, where $\mathcal{R}$ is the set of SynthSeg labels. Each voxel $q$ stores the label $\mathbf{S}^{(T+1)}(q)$ of its anatomical region. The label map is used only to construct the loss weights and is not provided to ProgFormer or used during inference.

Each label $r\in\mathcal{R}$ is assigned a predefined weight $\lambda_r$. Progression-relevant regions receive higher weights, while the remaining tissue and background retain lower weights. The weight of voxel $q$ is $w(q)=\lambda_{\mathbf{S}^{(T+1)}(q)}$. Let $\Omega$ denote the set of all $DHW$ voxels. The weighted flow-matching objective is
\begin{equation}
    \mathcal{L}_{\mathrm{FM}}
    =
    \mathbb{E}_{\mathbf{x}^{(T+1)},\boldsymbol{\epsilon},\tau}
    \left[
        \frac{1}{|\Omega|}
        \sum_{q\in\Omega}
        w(q)
        \left(
            \mathbf{\nu}_{\theta}(q)
            -
            \mathbf{\nu}^{*}(q)
        \right)^2
    \right].
    \label{eq:weighted_flow_loss}
\end{equation}

\paragraph{Inference.}
The inference procedure is shown in the right panel of Figure~\ref{fig:adaln_attention}. Generation starts from $\mathbf{y}_0\sim\mathcal{N}(\mathbf{0},\mathbf{I})$, while $\mathcal{H}_T$, $a^{(T+1)}$, and $\mathbf{c}$ remain fixed. For $K$ Euler steps, let $\Delta\tau=1/K$ and $\tau_k=k\Delta\tau$. At step $k$, ProgFormer predicts
\begin{equation}
    \mathbf{\nu}_k
    =
    \mathbf{\nu}_{\theta}
    \left(
        \mathbf{y}_k,\tau_k
        \mid
        \mathcal{H}_T,
        a^{(T+1)},
        \mathbf{c_{T+1}}
    \right),
    \label{eq:inference_velocity}
\end{equation}
and updates the current flow state as
\begin{equation}
    \mathbf{y}_{k+1}
    =
    \mathbf{y}_k
    +
    \Delta\tau\,\mathbf{\nu}_k.
    \label{eq:euler_sampling}
\end{equation}
The updated state $\mathbf{y}_{k+1}$ and flow time $\tau_{k+1}$ are then used in the next step. After $K$ updates, the final state is returned as the predicted future MRI, $\widehat{\mathbf{x}}^{(T+1)}=\mathbf{y}_K$.

\section{Experiments}
\label{sec:experiments}

\subsection{Experimental Setup}
\label{sec:setup}

\paragraph{Datasets.}

We evaluate ProgFormer on three longitudinal Alzheimer's disease MRI cohorts: ADNI~\cite{ADNI}, AIBL~\cite{AIBL}, and OASIS~\cite{OASIS}. All T1-weighted scans undergo N4 bias correction~\cite{N4Bias}, skull stripping, affine registration to the MNI152 ~\cite{MNI152}, resampling to $2.25mm^3$, and intensity normalization. Each cohort is split subject-wise into $80\%$, $5\%$, and $15\%$ training, validation, and test sets, respectively.

\paragraph{Metrics.}

Following prior work~\cite{BrLP,TADM3D,SADM,DeltaLFM}, we report peak signal-to-noise ratio (PSNR) and structural similarity (SSIM) to evaluate image quality. To assess regional structural fidelity, we report region-level mean absolute error (R-MAE) over clinically relevant structures, including the hippocampus, amygdala, lateral ventricles, thalamus, and cerebrospinal fluid. R-MAE averages the absolute differences between the predicted and target volumes of these structures, with lower values indicating greater regional accuracy. 

\paragraph{Comparison methods.}
We compare ProgFormer with BrLP~\cite{BrLP}, CounterSynth~\cite{countersynth}, TADM-3D~\cite{TADM3D}, and SADM~\cite{SADM}. All methods use the same preprocessing and input spatial dimensions. In Table~\ref{tab:main_results}, $I$ denotes the number of scans provided to each method. For SADM, we evaluate its single-input configuration ($I=1$) and additionally train and evaluate a multiple-input configuration using all available observed scans ($I=T$) when $T>1$.

\paragraph{Implementation details.}

ProgFormer uses $L_c=8$ coarse blocks and $L_f=4$ fine blocks. The coarse and fine hidden dimensions are $d_c=384$ and $d_f=256$, respectively. We use patch size $p=4$, producing $N=9600$ patch tokens and $P=64$ voxel tokens per patch, with six attention heads. The interleaved variant uses $N_{\mathrm{stage}}=4$ stages. Training uses AdamW with learning rate $10^{-4}$, weight decay $10^{-2}$, batch size $4$ per GPU, and gradient clipping at norm $1.0$. Inference uses $K=100$ Euler steps. All experiments are conducted on $8 \times$ NVIDIA A100 $80$GB GPUs.

\begin{figure*}[!ht]
\centering
\includegraphics[width=\linewidth]{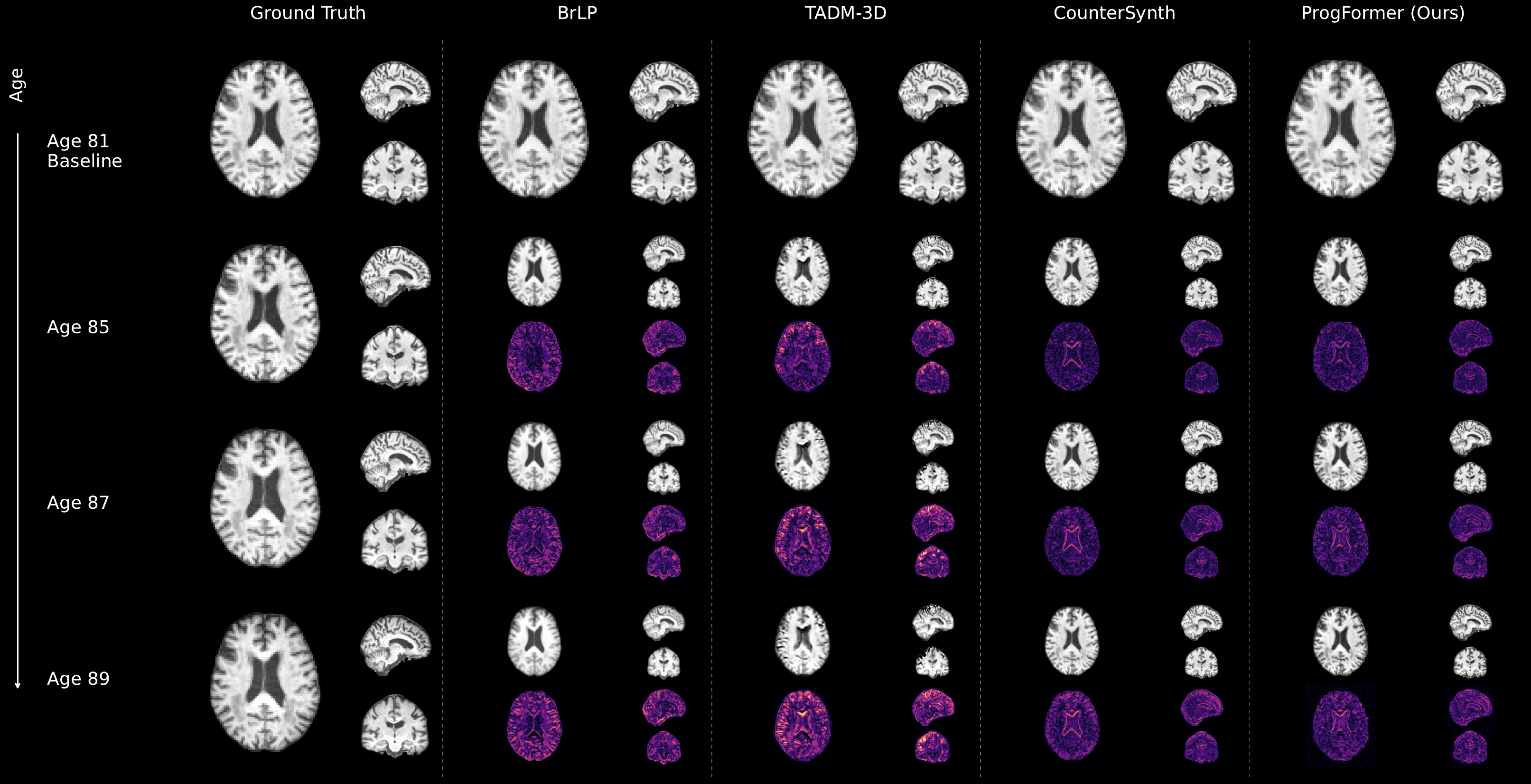}
\caption{Longitudinal prediction for an ADNI subject.
The first column shows the ground-truth follow-up scans. Columns 2--5 show predictions from BrLP, TADM-3D,  CounterSynth, and our ProgFormer, respectively. Starting from the observed scan at age $81$, each method predicts the follow-up scans at ages $85$, $87$, and $89$. Each prediction is shown together with its voxel-wise absolute-error map relative to the corresponding ground-truth scan. 
The brighter color denotes larger error.}
\label{fig:progression}
\end{figure*}

\begin{table}[!ht]
\centering
\caption{Component ablations of ProgFormer-S on ADNI. Best results are in \textbf{bold}, and second-best results are \underline{underlined}. SSIM is presented on a $0$--$100$ scale.}
\label{tab:ablation}
\small
\renewcommand{\arraystretch}{1.10}
\setlength{\tabcolsep}{4pt}
\begin{tabular*}{\columnwidth}{@{\extracolsep{\fill}}lcc@{}}
\toprule
\textbf{Variant} & \textbf{PSNR} $\uparrow$ & \textbf{SSIM} $\uparrow$ \\
\midrule
\multicolumn{3}{l}{\textit{Pairwise input} (${I}=1$)} \\
\midrule
w/o Coarse Pathway
& $7.23 \pm 0.19$
& $0.28 \pm 0.22$ \\

w/o Fine Pathway
& ${24.79 \pm 1.27}$
& $78.19 \pm 11.01$ \\

w/o Cross-Attn.
& ${24.79 \pm 1.48}$
& ${91.42 \pm 1.87}$ \\

w/o Region Weights
& $24.75 \pm 1.19$
& $87.60 \pm 1.66$ \\

\textbf{ProgFormer-S (full)}
& \bestval{26.39 \pm 1.68}
& \bestval{93.56 \pm 2.12} \\

\midrule
\multicolumn{3}{l}{\textit{Trajectory input} (${I}=T, 1<T\leq4$)} \\
\midrule
w/o Coarse Pathway
& $7.63 \pm 0.17$
& $0.32 \pm 0.21$ \\

w/o Fine Pathway
& ${26.49 \pm 1.59}$
& $83.79 \pm 5.28$ \\

w/o Cross-Attn.
& $26.13 \pm 1.56$
& $87.69 \pm 1.70$ \\

w/o Region Weights
& $24.45 \pm 1.13$
& ${88.81 \pm 1.40}$ \\

\textbf{ProgFormer-S (full)}
& \bestval{27.90 \pm 1.63}
& \bestval{96.17 \pm 2.04} \\
\bottomrule
\end{tabular*}

\vspace{2pt}
\begin{minipage}{\columnwidth}
\footnotesize
Cross-Attn. denotes coarse-to-fine cross-attention. 
\end{minipage}
\vspace{-0.5cm}
\end{table}

\subsection{Results}
\label{sec:main_results}

\paragraph{Pairwise prediction.}

As shown in Table~\ref{tab:main_results},
ProgFormer-S achieves the best PSNR on all three cohorts, with $26.39$ on ADNI, $26.95$ on AIBL, and $26.32$ on OASIS. It outperforms the strongest competing result by $0.20$, $0.22$, and $0.28$, respectively, and also achieves the best SSIM on AIBL and OASIS. These results show that sequential coarse modelling and fine refinement provide consistently strong image fidelity across cohorts.

ProgFormer-I performs best on regional accuracy, achieving an R-MAE of $0.230$ on ADNI, $0.223$ on AIBL, and $0.281$ on OASIS,
while also attaining the best SSIM on ADNI. 

The sequential and interleaved variants thus offer different advantages. ProgFormer-S is more consistent on image-level metrics, whereas ProgFormer-I better preserves regional structure. We use ProgFormer-S as the primary configuration in the remaining experiments.



\paragraph{Trajectory prediction.}

Pairwise and ${I}=1$ trajectory prediction both use a single observed scan, but differ in how observation-target pairs are selected. Their results are consequently similar. The trajectory protocol uses the final visit as the target, providing a controlled comparison between systems using only the latest scan and systems using the observed history.


Compared with the strongest ${I}=1$ result for each metric, using the observed history improves PSNR by $1.28$ on ADNI, $0.17$ on AIBL, and $0.79$ on OASIS. It improves SSIM by $0.37$ on ADNI and $0.20$ on AIBL, and reduces R-MAE by $0.044$ and $0.013$, respectively. On OASIS, the history model has $0.06$ lower SSIM and $0.002$ higher R-MAE. The history input consistently benefits image fidelity, while its effect on regional accuracy depends on the cohort.

ProgFormer-S also outperforms SADM when both methods use multiple observed scans. Under the same ${I}=T$ setting, it improves PSNR by $0.58$ on ADNI, $2.80$ on AIBL, and $1.96$ on OASIS, while reducing R-MAE by $0.163$, $0.049$, and $0.026$, respectively. These results show the benefit of integrating longitudinal information at the patch level before transferring it to voxel-level refinement.

Figure~\ref{fig:progression} shows sequential predictions from the observed scan at age $81$ to ages $85$, $87$, and $89$. For later predictions, the pairwise baselines use their most recent prediction, whereas ProgFormer uses the baseline scan and its earlier predictions. Ground-truth follow-ups are used only for visualization and error calculation. Compared with the accumulating errors of the baselines, ProgFormer preserves the ventricular configuration and cortical outline more consistently, with smaller and more localized errors across the three target ages.


\begin{figure}[ht]
\centering
\includegraphics[width=\linewidth]{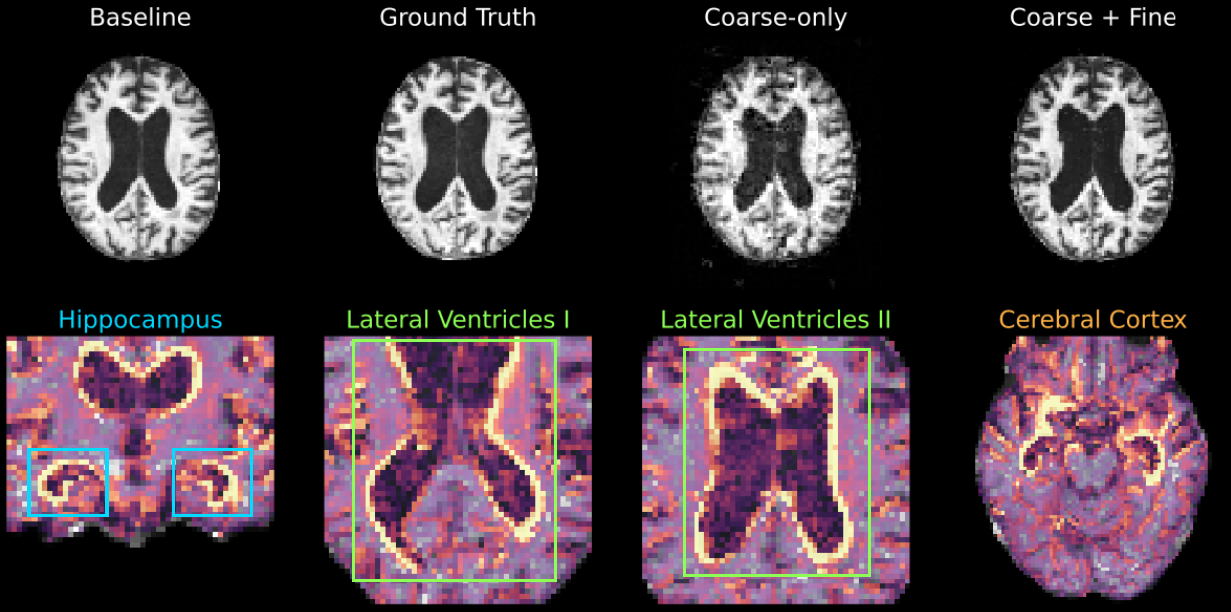}
\caption{Coarse-to-fine refinement.
The top row shows the baseline scan, ground-truth follow-up, coarse-only prediction, and full coarse-to-fine prediction. The bottom row shows the absolute voxel-wise difference between the two predictions in selected brain regions. Brighter colours indicate larger modifications introduced by the fine pathway, which are concentrated around hippocampal, ventricular, and cortical boundaries. 
}
\label{fig:coarse_fine}
\vspace{-0.4cm}
\end{figure}

\subsection{Ablation Study}
\label{sec:ablation}

Table~\ref{tab:ablation} validates the roles of the two pathways and their interaction. Removing the coarse pathway causes prediction to collapse, reducing PSNR from $26.39$ to $7.23$ under pairwise input and from $27.90$ to $7.63$ under trajectory input. This confirms that patch-level reasoning establishes the global brain structure required for prediction.

Removing the fine pathway decreases PSNR by $1.60$ and $1.41$ dB under pairwise and trajectory inputs, respectively, while SSIM decreases by $15.37$ and $12.38$ points. Removing coarse-to-fine attention decreases PSNR by $1.60$ dB under pairwise input and $1.77$ dB under trajectory input. The corresponding SSIM reductions are $2.14$ and $8.48$ points. The greater reduction under trajectory input indicates that transferring coarse features becomes particularly important when they contain information aggregated from several observed scans. Voxel processing is therefore most effective when grounded in the patch-level structural and longitudinal representation.

Removing the regional weights decreases PSNR by $1.64$ and $3.45$ dB and SSIM by $5.96$ and $7.36$ points under pairwise and trajectory inputs, respectively. The consistent degradation indicates that emphasizing progression-relevant regions improves overall prediction quality despite stable tissue and background occupying most of the scan.


Figure~\ref{fig:coarse_fine} visualizes the voxel-level refinement made by the fine pathway. The coarse-only prediction establishes the overall brain structure, whereas the full prediction more closely reproduces local boundaries in the ground-truth scan. The strongest refinements occur around the hippocampus, lateral ventricles, and cerebral cortex, while more stable regions receive smaller updates. This concentration in progression-relevant regions suggests that the fine pathway primarily captures localized longitudinal changes rather than reconstructing the complete scan.

\section{Conclusion}
\label{sec:conclusion}

We propose ProgFormer, an end-to-end hierarchical voxel-space Diffusion Transformer for longitudinal brain MRI prediction. ProgFormer utilizes patch tokens for  spatial and longitudinal reasoning and voxel tokens for local refinement, with coarse-to-fine attention providing the coarse representations as spatiotemporal grounding. Across ADNI, AIBL, and OASIS, the sequential variant achieves the highest pairwise PSNR on all three cohorts, while the interleaved variant achieves the lowest pairwise regional error. Under trajectory evaluation, the sequential model using the observed history achieves the highest PSNR across all three cohorts and the strongest image and regional performance on ADNI and AIBL. The ablations and visualisations further show that the coarse pathway provides the primary volumetric prediction, while the fine pathway uses the coarse information to recover localised structural details. These results demonstrate hierarchical voxel-space modelling as an effective approach to longitudinal volumetric prediction.

\vspace{1em}
\section*{Acknowledgement}
This research was supported by the NVIDIA Academic Grant Program using NVIDIA A100 (80GB) Tensor Core GPUs accessed via the NVIDIA Brev cloud platform.

\bibliography{aaai2027}

\end{document}